\documentclass[lettersize,journal]{IEEEtran}
\usepackage{amsmath,amsfonts,amssymb}
\usepackage{algorithmic}
\usepackage{algorithm}
\usepackage{array}
\usepackage{textcomp}
\usepackage{stfloats}
\usepackage{url}
\usepackage{verbatim}
\usepackage{booktabs}
\usepackage{mdwmath}
\usepackage{mdwtab}
\usepackage{eqparbox}
\usepackage{multirow}
\usepackage{xcolor}
\usepackage{makecell}
\usepackage{enumerate}
\usepackage{array}
\usepackage{graphicx}
\usepackage{subfigure}

\begin{document}

\title{Multimodal Hyperspectral Image Classification via Interconnected Fusion}

\author{Lu~Huo, Jiahao~Xia, Leijie~Zhang, Haimin~Zhang, Min~Xu \\Faculty of Engineering and IT, University of Technology Sydney

\thanks{This paper was produced by the IEEE Publication Technology Group. They are in Piscataway, NJ.}
}

\markboth{Journal of \LaTeX\ Class Files,~Vol.~14, No.~8, August~2021}%
{Shell \MakeLowercase{\textit{et al.}}: A Sample Article Using IEEEtran.cls for IEEE Journals}


\maketitle

\begin{abstract}
Existing multiple modality fusion methods, such as concatenation, summation, and encoder-decoder-based fusion, have recently been employed to combine modality characteristics of Hyperspectral Image (HSI) and Light Detection And Ranging (LiDAR). However, these methods consider the relationship of HSI-LiDAR signals from limited perspectives. More specifically, they overlook the contextual information across modalities of HSI and LiDAR and the intra-modality characteristics of LiDAR. In this paper, we provide a new insight into feature fusion to explore the relationships across HSI and LiDAR modalities comprehensively. An Interconnected Fusion (IF) framework is proposed. Firstly, the center patch of the HSI input is extracted and replicated to the size of the HSI input. Then, nine different perspectives in the fusion matrix are generated by calculating self-attention and cross-attention among the replicated center patch, HSI input, and corresponding LiDAR input. In this way, the intra- and inter-modality characteristics can be fully exploited, and contextual information is considered in both intra-modality and inter-modality manner.  
These nine interrelated elements in the fusion matrix can complement each other and eliminate biases, which can generate a multi-modality representation for classification accurately. Extensive experiments have been conducted on three widely used datasets: Trento, MUUFL, and Houston. The IF framework achieves state-of-the-art results on these datasets compared to existing approaches.

\end{abstract}


\section{Introduction}
\begin{figure}[htbp]
  \begin{center}
  \includegraphics[width=3.5in]{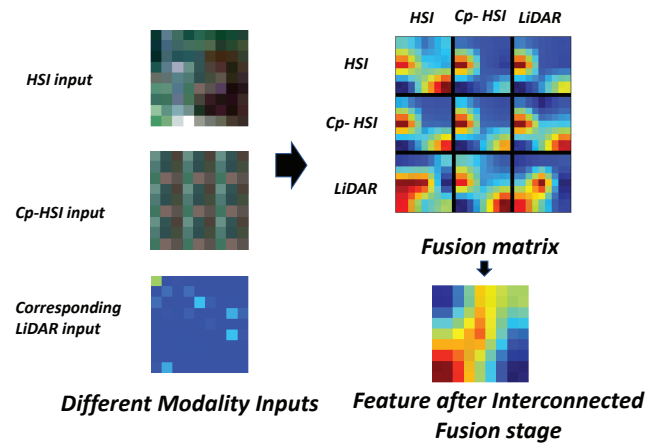}\\
  \caption{The grass sample consists of three inputs, including the HSI, the center patch of the HSI (cp-HSI), and the corresponding LiDAR signals. The green part in false color images of the HSI signal presents the grass part while the brown color shows the soil area. These three inputs can generate nine interconnected perspectives for HSI-LiDAR signals. Moreover, these nine interrelated views can complement each other and work together to generate the integrated feature, which can eliminate the bias from each perspective and focus on the grass area.  }\label{circuit_diagram}
  \end{center}
\end{figure}


Hyperspectral image (HSI) contains enormous spectral information in each pixel \cite{7565539,shi2019multiscale}, which enables the extraction of fine features to identify the material on the earth's surface \cite{bajcsy2004methodology,974718,8738051}. However, lacking altitude information, HSI has hardly been used to identify objects with occlusions, such as clouds. Using HSI data to differentiate the same object from similar categories, such as grass on the roof or grass on the ground, is impossible due to different altitudes\cite{hong2020more, li2022deep}.
Capturing altitude information, Synthetic Aperture Radar (SAR) \cite{chang2004data,7001095} or Light Detection And Ranging (LiDAR) \cite{6776408} collaborating with HSI signal are used to detect objects with occlusions or even totally hidden behind obstacles. Thus, the integration of different modalities has attracted increasing research attention in the field of remote sensing.



Most multi-modality fusion methods encode feature representations for one modality and then fuse the features of multiple modalities for classification \cite{xu2017multisource, hong2020deep, zhang2021information}. Traditional feature fusion approaches, such as concatenation, summation, and encoder-decoder-based fusion, consider feature fusion from limited viewpoints, which cannot consider the inter-modality relationship comprehensively \cite{xu2017multisource, hong2020deep, zhang2021information, wu2021convolutional, wang2022multi}. After feature concatenation, the fused neurons from the corresponding intra-modality feature are triggered while neurons from another intra-modality feature are inhibited \cite{hong2020more}. Most HSI-LiDAR fusion methods only consider the spatial correlation within a single modality, such as an HSI signal, while neglecting the contextual information among image patches from inter-modality, limiting the final performance. Besides, some of these fusion methods concentrate on the intra-modality relationship within HSI and overlook that in LiDAR, which leads to information deficiency \cite{xu2022multimodal}. In addition, existing feature fusion approaches lack generalization ability for unseen training samples and are prone to overfitting \cite{wang2020makes}.

In this paper, we propose an Interconnected Fusion (IF) framework to enrich intra-modality and inter-modality information and obtain a satisfactory data representation, as shown in Fig. 1. 
We divide HSI input into $3\times3$ patches, extract the center patch of HSI input, and replicate it to the size of the HSI input as an extra input to our framework.  
Then, a fusion matrix is introduced to capture and represent the interconnected correlations among three multi-modality inputs, including HSI, the replicated center patch of HSI, and the corresponding LiDAR through applying self-attention and cross-attention operations. Later, nine interrelated elements in the fusion matrix can work together to accurately generate an integrated feature representation through the feature fusion layer. 

 Overall, our main contributions are summarised as follows:
\begin{enumerate}[1)]
\item Extracting the center patch of HSI as an extra input to our framework, the contextual information between the center patch and nearby patches can be extracted to enrich the information from intra- and inter-modality.  

\item Different from traditional fusion methods such as concatenation, summation, and encoder-decoder-based fusion, the nine interconnected elements in the fusion matrix are first introduced as a feature map for HSI-LiDAR signals, which can complement each other and work together to eliminate the bias from each element.   




\item Compared to other methods, the proposed method outperforms existing methods on three widely used datasets, including Trento, MUUFL, and Houston. In addition, we extend our experiment to study the effects of the change in patch size and fusion strategies, including early, middle, and late. We found that the IF framework can achieve robust performance on different fusion strategies when the input patch has more pixels, which means IF is a robust framework that achieves similar performance in different fusion strategies.



\end{enumerate}

\section{Related work}
\subsection{Approaches of single modality signal representation}
The commonly used single modality signal representation approaches in the field of HSI are Convolutional Neural Networks (CNN) based methods \cite{li2016hyperspectral, xu2017multisource, lee2017going, hong2020deep, hang2020classification, zhang2021information}. Deep CNN with pixel-pair features (CNN-PPF) \cite{li2016hyperspectral} and 1D-CNN \cite{chen2016deep} with $\ell_2$ regularization employ pixel-pair CNN structure.  
However, pixel-wise information for the HSI area only explores the spectral feature at the pixel level without any spatial relationship. The spectral data of nearby pixels have similar attributes 
which is suitable to explore more discrimination power for HSI data \cite{he2017recent,ghamisi2018new, zhong2017spectral, 9460777, 9552385,plaza2009recent, 8314827,lee2017going,ji2013spectral}. However, suppose the CNN model extracts spectral and spatial information together. In that case, they will pay considerable attention to spatial direction knowledge rather than spectral information, which means a significant amount of hyperspectral information on each pixel is not fully leveraged  \cite{hong2021spectralformer}. Moreover, the CNN-based models have many limitations, such as neglecting long-range dependencies and overlooking the positional information \cite{amir2021deep}.

The Vision Transformer (ViT) \cite{dosovitskiy2020image} achieves better performance than the CNN structure. Moreover, transformer-based models provide more robustness and generalizable features than CNN-based models with fewer parameters and a lower risk of overfitting \cite{naseer2021intriguing, zhou2021convnets}. SpectralFormer \cite{hong2021spectralformer} can learn group-wise spectral embeddings from neighboring bands of HSI. Furthermore, some transformer-based methods employ the spectral-spatial transformer architecture to get spatial and spectral attention together \cite{zhong2021spectral, sun2022spectral}. Unfortunately, the parameters of most transformer-based models are enormous and computationally costly \cite{han2022survey}. In addition, the performance of these models relies heavily on large-scale training data \cite{dosovitskiy2020image}.  Thus, many challenges remain in training transformer-based models. In our framework, we employ ViT as the backbone to capture general dense visual features \cite{amir2021deep} coupled with the neighborhood relationship of the center patch and nearby patches to extract the context-related feature. To alleviate the data-hungry limitation of ViT, the depth of our framework is three.

\subsection{Approaches of multiple modality fusion}

The feature fusion algorithms in multi-modal HSI fields usually employ traditional fusion methods, including summation, concatenation, encoder-decoder, and cross-fusion \cite{xu2017multisource, hong2020deep, zhang2021information, wu2021convolutional, wang2022multi}. Two-branch CNN employs the two sub-network to extract the features from HSI and LiDAR data respectively and then combine these two different features using concatenation \cite{xu2017multisource}. In addition, the weight sharing for two branch CNNs (Co-CNN) has been applied for HSI and LiDAR feature extraction. After that, these two types of features are fused at the feature level using summation or maximization strategy \cite{hang2020classification}. 
Encoder–decoder architecture (EndNet) has been employed in this field. They use an autoencoder to integrate and reconstruct the different modalities' information \cite{hong2020deep}. Another autoencoder-based model is the Interleaving Perception Convolutional Neural Network (IP-CNN), which employs a bidirectional autoencoder for information fusion, feature extraction, and reconstruction for HSI and LiDAR data sources \cite{zhang2021information}. Moreover, a shared and specific feature learning (S2FL) model divides the multi-modal HSI data into two broad types, including shared space for different modalities and specific space for each modality \cite{hong2021multimodal}. Furthermore, the MDL-RS framework employs a compactness-based cross-fusion to get a more diversified relationship from different modalities \cite{hong2020more}. However, the feature extracted by CNN structure has a substantial prejudice towards texture and a troubling shortage of positional information \cite{amir2021deep}. Thereby, the CNN structure remains limited to capturing the global interaction of each element of the input.

Recently, ViT has been applied in HSI multi-model field because ViT can achieve competitive performance compared to CNN. The multimodal fusion transformer (MFT) network combines the characteristics of different modalities through cls token from another modality \cite{roy2022multimodal}. Nevertheless, these fusion methods overlook the contextual information for multi-modal signals and the information of the modality itself, which will retain limited performance while our framework extracts intra- and inter-modality with the contextual information between the center patch and the nearby patches.

\section{Interconnected Fusion Framework}
\begin{figure*}[htbp]
  \begin{center}
  \includegraphics[width=7in]{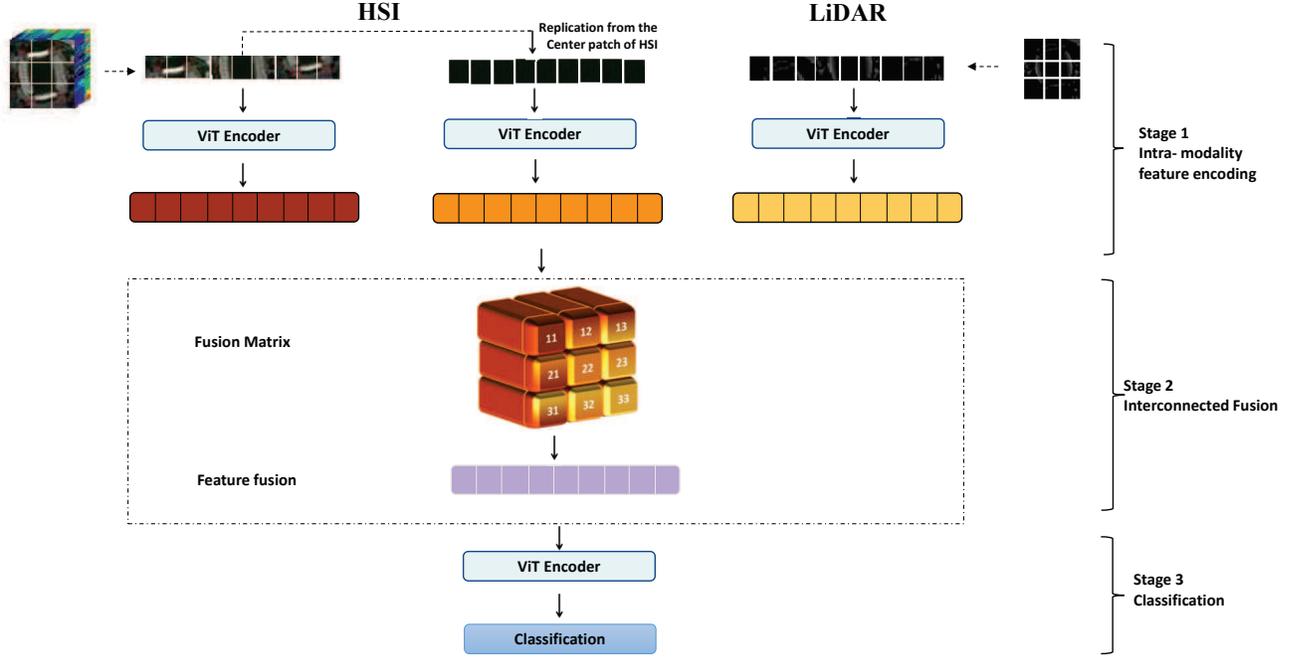}\\
  \caption{The proposed IF framework contains three stages: Intra-modality feature encoding, interconnected fusion, and classification. For Intra-modality feature encoding, the center patch of HSI is introduced as an extra input to explore the contextual information between the center patch and nearby patches for intra- and inter-modality. Regarding the interconnected fusion stage, the nine interconnected perspectives in the Fusion Matrix can represent the intra- and inter-modality with contextual information. These nine interrelated views can complement each other and eliminate the bias from each other through the feature fusion layer. For the final stage, the integrated feature can be calculated, and the classification result can be obtained.}\label{circuit_diagram}
  \end{center}
\end{figure*}


The proposed IF framework contains three stages: intra-modality feature encoding, interconnected fusion, and classification (see Fig. 2 and Algorithm 1). The first stage employs ViT encoders to extract three intra-modality features for the replicated center patch of HSI input, HSI input, and LiDAR input separately. Then, the second stage introduces the fusion matrix to gather nine perspectives for intra- and inter-modality information through calculating self-attention and cross-attention of three intra-modality features. Afterward, the nine interrelated elements in the Fusion Matrix can complement each other and work as an integrated feature through the feature fusion layer.
Moreover, the last stage employs the ViT Encoder to get more abstract integrated features and classification results. 


\begin{algorithm}[h]
    \caption{The pseudocode for the core of an implementation of IF framework}
    \label{alg:4}
    \begin{algorithmic}[1]
    \STATE{\textbf{Initialization:} Set $t=0$, the input data $X_{h1}$, $X_{h2}$ from HSI data source and $X_{l}$ from LiDAR data source. Specifically, the center patch of the HSI input $X_{h2}$ is replicated to the same size as the HSI input  $X_{h1}$. }
        \FOR{$t=0,1,2,\cdots,$}
            \STATE {\textbf{Intra-modality feature encoding:} Extract feature representations  $Z^{l}_{h1}$, $Z^{l}_{h2}$ and $Z^{l}_{l}$ for the input data $X_{h1}$, $X_{h2}$, $X_{l}$ through ViT encoders separately (refer to Eq.(4)};\
            \STATE {\textbf{Interconnected Fusion:} 
           calculate \textbf{nine interconnected perspectives}:
           
$Z^{c}_{ij}= f_{views}(Z^{l}_1, Z^{l}_2, Z^{l}_3) \quad i\textsc{,}j\in\{1,2,3\}$;

define \textbf{fusion matrix}: 
$Z^{m} =  \begin{bmatrix} Z^{c}_{11} & Z^{c}_{12} & Z^{c}_{13} \\ Z^{c}_{21} & Z^{c}_{22} & Z^{c}_{23} \\  Z^{c}_{31} & Z^{c}_{32} & Z^{c}_{33}\end{bmatrix}$

calculate \textbf{an integrated and corrected feature} $\hat{Z^{f}}$ and then updated to $Z^{f}$(refer to Eq.(9) and Eq.(10)};\
            \STATE {\textbf{Classification:} Processing the combined feature $Z^{f}$ through the ViT Encoder (refer to Eqs.(11) and (12)) and getting classification result};\
        \ENDFOR
    \end{algorithmic}
\end{algorithm}

\subsection{Intra-modality feature encoding}


For intra-modality feature encoding, the center patch of HSI is introduced as extra input for further exploring the contextual information between the center patch and nearby patches.
The inputs contain three data sources, including HSI $X_{h1}$, the corresponding LiDAR $X_{l}$, and the center patch $X_{h2}$ of HSI data feeding to three separate encoders, $H, H_p$ for HSI and the $L$ for LiDAR. The input for HSI and LiDAR has the same patch numbers $p$ and the embedding dimensions $D$ after patch embedding. The data of $X_{h2}$ comes from the center patch of $X_{h1}$ duplicated $p$ times (same patch numbers with HSI and LiDAR sample). Then, the HSI and LiDAR samples are passed through the transformer layer including Multi-head Self-Attention (MSA), Feed-Forward Network (FFN), and Layer Normalization (LN) blocks. The MSA can be calculated by concatenating self-attention (SA) results for different heads and then projected by a weight matrix W (see Equations 1, 2): 
\begin{equation}\label{MSA}
\begin{aligned}
MSA(z) =  [SA_1(z), ..., SA_h(z)]W \quad \quad W\in \mathbb{R}^{hD_q\times D}
\end{aligned}
\end{equation}
where $h$ is the number of heads for MSA, and $SA(z)$ can be calculated:
\begin{equation}\label{attention}
\begin{aligned}
 SA(z) = softmax(\frac{qk^\intercal}{\sqrt{D_q}})v \quad \quad  &[q,k,v] = zU_{qkv}\\
 & U_{qkv}\in \mathbb{R}^{D\times3D_q}
\end{aligned}
\end{equation}
where q, k, and v mean query, key, and value, respectively.
The MSA results $\hat{Z^{l}_{h1}}$, $\hat{Z^{l}_{h2}}$ and $\hat{Z^{l}_{l}}$ can be calculated:
\begin{equation}\label{self-attention}
\begin{aligned}
\hat{Z^{l}_{i}} = MSA(LN(Z^{l-1}_{i})) + Z^{l-1}_{i} \quad \quad i\in \{h1,h2,l\}
\end{aligned}
\end{equation}
Then, the output of MSA can be updated by the FFN block. 
\begin{equation}\label{ffn}
Z^{l}_i = MLP(LN(\hat{Z^{l}_{i}})) + \hat{Z^{l}_{i}} \quad \quad i\in \{h1,h2,l\} \\
\end{equation}
Thus, three different results $Z^{l}_{h1}$, $Z^{l}_{h2}$ and $Z^{l}_{l}$ are generated from encoder $H, H_p$ and $L$, respectively. Furthermore, after Intra-Modality data processing, these three outputs have the same number of patches and dimensional embeddings.


\subsection{Interconnected fusion}

\begin{figure*}[htbp]
  \begin{center}
  \includegraphics[width=7in]{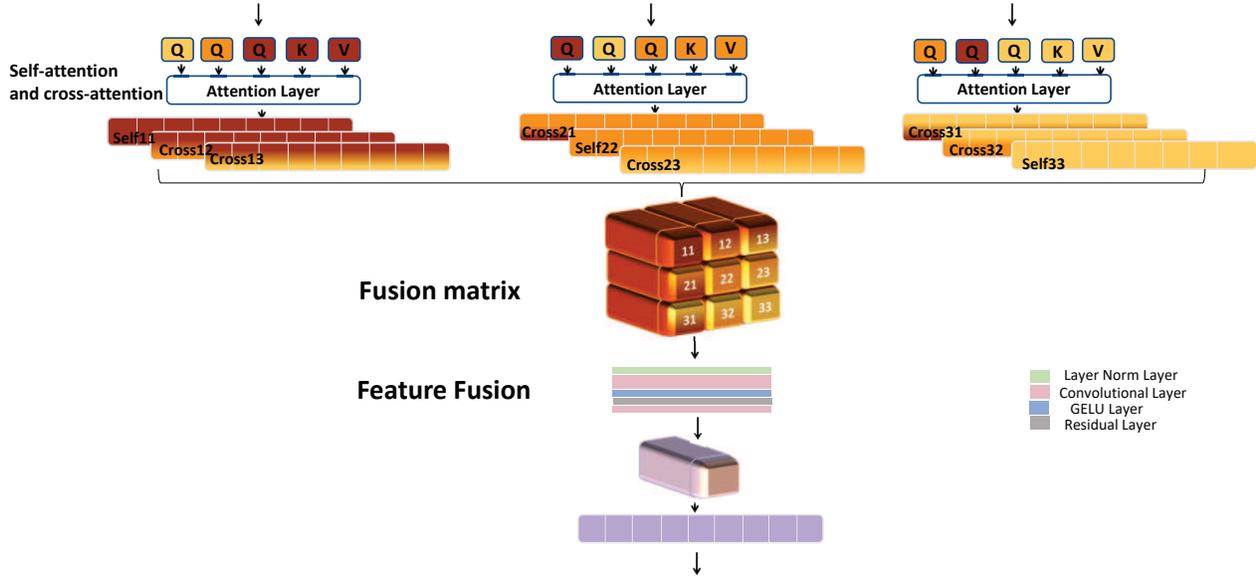}\\
  \caption{The interconnected fusion stage utilizes the
fusion matrix to gather nine perspectives for intra-modality and inter-modality information with the neighborhood correlation of the center patch and nearby patches. Then, the nine interrelated views in the Fusion Matrix can complement each other and work as an integrated feature through the Feature Fusion layer.
   }\label{circuit_diagram}
  \end{center}
\end{figure*}

The IF framework not only focuses on feature fusion from different modalities but also on leveraging comprehensive interactions from different modalities.
 Concerning the interconnected fusion stage, nine interconnected elements in a fusion matrix are proposed to represent the information from intra- and inter-modality and the contextual information across different modalities. The nine interrelated elements in the fusion matrix can complement each other and work as an integrated feature through the feature fusion layer. 
 
 First of all, the three intra-modality features $Z^{l}_{h1}$, $Z^{l}_{h2}$ and $Z^{l}_{l}$ are calculated in previous stage. Then, nine interconnected perspectives $Z^{c}_{ij}$ can be generated through self-attention and cross-attention of these three intra-modality features: 
\begin{equation}\label{fusion-block}
\begin{aligned}
Z^{c}_{ij} = f_{views}(Z^{l}_{h1},Z^{l}_{h2},Z^{l}_{l}) \quad \quad i\textsc{,}j\in\{1,2,3\} \\
\end{aligned}
\end{equation}
when $i=j$, the self-attention values $\hat{Z^{c}_{ij}}$ can be calculated by $k$, $q$, and $v$ generated from the same input $i$. On the contrary, when $i\not=j$, cross-attention values $\hat{Z^{c}_{ij}}$  can be computed by $k$ and $v$ from the same input $i$ while $q$ from another data source $j$:
\begin{equation}\label{cross-attention}
\begin{aligned}
\hat{Z^{c}_{ij}} = MSA(q_j, k_i, v_i) + Z^{l}_{i}
\quad \quad i\textsc{,}j\in\{1,2,3\}
\end{aligned}
\end{equation}
The value of $\hat{Z^{c}_{ij}}$ can be updated by residual connection and the FFN block:  
\begin{equation}\label{cross-attention}
\begin{aligned}
Z^{c}_{ij} = MLP(LN(\hat{Z^{c}_{ij}})) + \hat{Z^{c}_{ij}}  \quad \quad i\textsc{,}j\in\{1,2,3\}
\end{aligned}
\end{equation}
Then, the $3\times3$ fusion matrix can be defined using these nine views. 
\begin{equation}\label{fusion-matrix}
\begin{aligned}
Z^{m} = \begin{bmatrix} Z^{c}_{11} & Z^{c}_{12} & Z^{c}_{13} \\ Z^{c}_{21} & Z^{c}_{22} & Z^{c}_{23} \\  Z^{c}_{31} & Z^{c}_{32} & Z^{c}_{33}\end{bmatrix}
\end{aligned}
\end{equation}
In this fusion matrix, the intra-modality features consist of $Z^{c}_{11}$ for HSI, $Z^{c}_{33}$ for corresponding LiDAR, $Z^{c}_{22}$ for the center patch of HSI, and the HSI relationships between the center patch and nearby patches from two different views $Z^{c}_{12}, Z^{c}_{21}$, separately. Besides, $Z^{c}_{13},Z^{c}_{31},Z^{c}_{23},Z^{c}_{32}$ show the inter-modality correlations. More specifically, $Z^{c}_{13}$ and $Z^{c}_{31}$ represent the relationships of HSI and LiDAR patches at corresponding locations. In contrast, $Z^{c}_{23}$ and $Z^{c}_{32}$ provide the correlation across the center patch of the HSI and the nearby LiDAR patches. In addition, $Z^{c}_{12},Z^{c}_{21},Z^{c}_{23},Z^{c}_{32}$ shows the contextual information for intra and inter-modality both. Therefore, it is worth noting that nine interconnected elements in the fusion matrix contain the correlations between intra and inter-modality coupled with contextual information.

Each element of the fusion matrix contains bias because of limited information for multi-modal remote sensing signals. The feature fusion layer designed in the IF framework can effectively generate an integrated feature and eliminate the bias from distinct perspectives since these nine interrelated views in the fusion matrix can complement each other (the detailed discussion in Section 4 Ablation study).
Then, the $3\times3$ fusion matrix $Z^{m}$ can be compacted into a integrated feature $\hat{Z^{f}}$: 
\begin{equation}\label{FeatureFusion}
\begin{aligned}
\hat{Z^{f}} = f_{comp}(Z^{m}) 
\end{aligned}
\end{equation}
After that, this feature $\hat{Z^{f}}$ is updated by the FFN block and residual connection:
\begin{equation}\label{cross-attention}
\begin{aligned}
Z^{f} = \hat{Z^{f}} + FFN(\hat{Z^{f}})
\end{aligned}
\end{equation}
\subsection{Classification}
For the classification stage, the integrated result $Z^{f}$ is treated as a single feature and processed by the ViT encoder: 
\begin{equation}\label{Unity layer}
\hat{Z^{u}} = MSA(LN(Z^{f})) + Z^{f} 
\end{equation}
Moreover, $\hat{Z^{u}}$ is generated as a result of the MSA block and then processed by the FFN block:
\begin{equation}\label{Unity layer}
Z^{u} = MLP(LN(\hat{Z^{u}})) + \hat{Z^{u}}
\end{equation}
Finally, the result $Z^{u}$ can be calculated. The final classification result can be calculated from the $cls$ token.

\section{Experiments}



This section employs three well-known HSI multi-modal datasets to evaluate the performance of the IF framework. In addition, we compare the performance under different fusion strategies and patches sizes. We study different ablation experiments on the IF framework. 

\subsection{Datasets}

Houston 2013 Dataset was captured over the University of Houston campus and the nearby urban area in the 2013 IEEE GRSS Data Fusion Contest. This dataset includes HSI and LiDAR-derived Digital Surface Model (DSM) signals. The size of this dataset is 349 by 1905, with a spatial resolution of 2.5m. The HSI signal contains 144 bands from 0.38 to 1.05 $\mu$m. Moreover, this dataset consists of fifteen classes, including Grass Healthy, Grass Stressed, Grass Synthetic, Tree, Soil, Water, Residential, Commercial, Road, Highway, Railway, Parking Lot 1, Parking Lot 2, Tennis Court, and Running Track. The training and testing settings for each class are shown in Table I.

\begin{table}[]
\centering
\caption{The standard training and testing sets for each category on Houston Dataset}
\label{table3}
\begin{tabular}{ll|ll}
\hline
\bottomrule
\multicolumn{2}{c|}{Class}    & \multicolumn{2}{l}{Number of Samples}    \\ \hline
\multicolumn{1}{c|}{No.}  & Name  & \multicolumn{1}{c|}{Train}  & Test  \\ \hline
\multicolumn{1}{c|}{1} & Health grass  & \multicolumn{1}{c|}{198} & 1251 \\ \hline
\multicolumn{1}{c|}{2} & Stressed grass  & \multicolumn{1}{c|}{190} & 1254 \\ \hline
\multicolumn{1}{c|}{3} & Synthetic grass  & \multicolumn{1}{c|}{192} & 697 \\ \hline
\multicolumn{1}{c|}{4} & Tress  & \multicolumn{1}{c|}{188} & 1244 \\ \hline
\multicolumn{1}{c|}{5} & Soil  & \multicolumn{1}{c|}{186} & 1242  \\ \hline
\multicolumn{1}{c|}{6} & Water  & \multicolumn{1}{c|}{182} & 325  \\ \hline
\multicolumn{1}{c|}{7} & Residential  & \multicolumn{1}{c|}{196} & 1268  \\ \hline
\multicolumn{1}{c|}{8} & Commercial & \multicolumn{1}{c|}{191} & 1244  \\ \hline
\multicolumn{1}{c|}{9} & Road & \multicolumn{1}{c|}{193} & 1252 \\ \hline
\multicolumn{1}{c|}{10} & Highway  & \multicolumn{1}{c|}{191} & 1227  \\ \hline
\multicolumn{1}{c|}{11} & Railway & \multicolumn{1}{c|}{181} & 1235 \\ \hline
\multicolumn{1}{c|}{12} & Parking lot 1  & \multicolumn{1}{c|}{192} & 1233  \\ \hline
\multicolumn{1}{c|}{13} & Parking lot 2 & \multicolumn{1}{c|}{184} & 469 \\ \hline
\multicolumn{1}{c|}{14} & Tennis court & \multicolumn{1}{c|}{181} & 428 \\ \hline
\multicolumn{1}{c|}{15} & Running track & \multicolumn{1}{c|}{187} & 660  \\ \hline
\multicolumn{2}{c|}{Total}    & \multicolumn{1}{c|}{2832}  & 15029 \\ \hline
\bottomrule
\end{tabular}
\end{table}

Trento Dataset was measured in the rural area of Trento, Italy. The captured size is 600 by 166 pixels, and the spatial resolution is 1m. Furthermore, this dataset contains DSM and HSI signals with six distinct classes: Buildings, Woods, Apple trees, Roads, Vineyard, and Ground. The HSI contains 63 bands from 0.40 to 0.98 $\mu$m. The training and testing settings for each class are shown in Table II.

\begin{table}[]
\centering
\caption{The standard training and testing sets for each category on Trento Dataset.}
\label{table3}
\begin{tabular}{ll|ll}
\hline
\bottomrule
\multicolumn{2}{c|}{Class}    & \multicolumn{2}{l}{Number of Samples}    \\ \hline
\multicolumn{1}{c|}{No.}  & Name  & \multicolumn{1}{c|}{Train}  & Test  \\ \hline
\multicolumn{1}{c|}{1} & Apple trees & \multicolumn{1}{c|}{129}  & 4034  \\ \hline
\multicolumn{1}{c|}{2} & Buildings & \multicolumn{1}{c|}{125} & 2903  \\ \hline
\multicolumn{1}{c|}{3} & Ground  & \multicolumn{1}{c|}{105} & 479 \\ \hline
\multicolumn{1}{c|}{4} & Woods  & \multicolumn{1}{c|}{154} & 9123 \\ \hline
\multicolumn{1}{c|}{5} & Vineyard  & \multicolumn{1}{c|}{184} & 10501 \\ \hline
\multicolumn{1}{c|}{6} & Roads  & \multicolumn{1}{c|}{122} & 3174 \\ \hline
\multicolumn{2}{c|}{Total}    & \multicolumn{1}{c|}{819}  & 30214 \\ \hline
\bottomrule
\end{tabular}
\end{table}

The MUUFL dataset \cite{gader2013muufl} was measured over the University of Southern Mississippi campus in 2010. The size of this dataset is 325 by 220 pixels with 72 bands for the HSI signal. This dataset also contains two different LiDAR signals with different height. In our experiment, we concatenate these two LiDAR signals along the pixel location. Moreover, it contains 11 classes: Trees, Mostly grass, Mixed ground surface, Dirt and Sand, Road, Water, Building Shadow, Building, Sidewalk, Yellow curb, and Cloth panels. The training and testing settings for each class are shown in Table III.

\begin{table}[]
\centering
\caption{The standard training and testing sets for each category on MUUFL Dataset}
\label{table2}
\begin{tabular}{ll|ll}
\hline
\bottomrule
\multicolumn{2}{c|}{Class}    & \multicolumn{2}{l}{Number of Samples}    \\ \hline
\multicolumn{1}{c|}{No.}  & Name  & \multicolumn{1}{c|}{Train}  & Test  \\ \hline
\multicolumn{1}{c|}{1} & Trees  & \multicolumn{1}{c|}{150}  & 23246  \\ \hline
\multicolumn{1}{c|}{2}  & Mostly grass  & \multicolumn{1}{c|}{150} & 4270  \\ \hline
\multicolumn{1}{c|}{3}  & Mixed ground surface & \multicolumn{1}{c|}{150}  & 6882  \\ \hline
\multicolumn{1}{c|}{4} & Dirt and sand & \multicolumn{1}{c|}{150} & 1826 \\ \hline
\multicolumn{1}{c|}{5}  & Road & \multicolumn{1}{c|}{150} & 6687  \\ \hline
\multicolumn{1}{c|}{6} & Water  & \multicolumn{1}{c|}{150} & 466 \\ \hline
\multicolumn{1}{c|}{7}  & Building Shadow & \multicolumn{1}{c|}{150} & 2233 \\ \hline
\multicolumn{1}{c|}{8} & Building & \multicolumn{1}{c|}{150} & 6240 \\ \hline
\multicolumn{1}{c|}{9}  & Sidewalk & \multicolumn{1}{c|}{150} & 1385 \\ \hline
\multicolumn{1}{c|}{10}  & Yellow curb & \multicolumn{1}{c|}{150} & 183 \\ \hline
\multicolumn{1}{c|}{11}  & Cloth panels & \multicolumn{1}{c|}{150} &  269 \\ \hline
\multicolumn{2}{c|}{Total}    & \multicolumn{1}{c|}{1650}  & 53687 \\ \hline
\bottomrule
\end{tabular}
\end{table}

\subsection{Implementation details}
We run our model on a Linux system with Intel(R) Xeon(R) Gold 6132 CPU and Nvidia Quadro P5000 GPU. We implement three fusion strategies setting -- Early, Middle, and Late on the IF framework. If the number of total layers is N and Stage 1's layer number is M, the depth of Stage 3 is N-M-1 (see Fig. 4). To alleviate the data-hungry limitation of ViT, the total depth of our framework is three.
For the early fusion setting, the depth of Stage 1 is zero, and the layer number of Stage 3 is two. In contrast, for the late fusion setting, the depth of Stage 1 is two, and the layer number of Stage 3 is zero. For the middle fusion setting, the depth of Stage 1 is one, and the layer number of Stage 3 is one. Thus, the setting of fusion strategies for the IF framework can be customized by changing the depth of Stages 1 and 3.
 Furthermore, we take inspiration from ConvNet \cite{liu2022convnet} to extract the integrated feature effectively. The detailed setting of the feature fusion layer is shown in table IV. Thus, we can get an integrated feature $\hat{Z^{f}}$ after the feature fusion layer.

\begin{figure}[h]
  \begin{center}
  \includegraphics[width=3in]{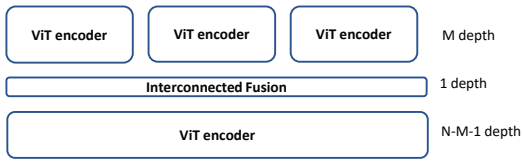}\\
  \caption{If the number of total layers is N and Stage 1's layer number is M, the depth of Stage 3 is N-M-1. }\label{circuit_diagram}
  \end{center}
\end{figure}


\begin{table}[]
\newcommand{\tabincell}[2]{\begin{tabular}{@{}#1@{}}#2\end{tabular}} 
\label{table8}
\centering
\caption{The detailed information for Feature Fusion layer}
\begin{tabular}{c|c|c}
\hline
\bottomrule
\tabincell{c}{Layer \\No.} & Layer type & Detailed setting \\ \hline
1 & Layer Norm & Shape [3,3] \\ \hline
2 & Convolutional layer & kernel size: (3,3); stride: (1,1); padding: 1\\ \hline
3 & Convolutional layer  & kernel size: (1,1); stride: (1,1); padding: 0\\ \hline
4 & Activate function & GELU()\\ \hline
5 & Residual layer & --\\ \hline
6 & Convolutional layer & kernel size: (3,3); stride: (1,1); padding: 0\\ \hline
 \bottomrule
\end{tabular}
\end{table}

\subsection{Result comparisons}

We compare the performance of our method with previous methods -- CNN-PPF \cite{li2016hyperspectral}, Two-branch CNN \cite{xu2017multisource}, Context CNN \cite{lee2017going}, EndNet\cite{hong2020deep}, MDL-RS\cite{hong2020more}, Co-CNN \cite{hang2020classification}, S2FL \cite{hong2021multimodal},
Spectralformer \cite{hong2021spectralformer}, MFT \cite{roy2022multimodal}
IP-CNN \cite{zhang2021information} in Table V, VI, and VII. Within these tables, it is noteworthy that the red values denote the best outcome in the present measurement, whereas the blue values indicate the second-best outcome.

\textbf{Evaluation on the Houston dataset} As shown in Table V, IF achieved the best performance. The overall accuracy (OA), the average accuracy (AA), and the Kappa coefficient of IF are more advantageous than the second-best method (IP-CNN) at around $5\%$, $3\%$, and $6\%$, respectively. Moreover, for most categories, the IF framework can get competitive results, while IP-CNN achieves the second-best performance, which means IF can gather comprehensive relationships for different modalities without the risk of data leakage like IP-CNN. The implementation of IF results in a notable enhancement of performance by $14\%$, which suggests that our proposed method is capable of effectively exploring a greater quantity of information in comparison to a single HSI modality. Compared with Two-branch CNN, context CNN, Co-CNN, EndNet, MDL-RS, and S2FL, the result of IF improves $8\%$ since 
the collection of nine distinct views for HSI-LiDAR signals is found to encompass a wealth of informative details to both intra- and inter-modality relationships, inclusive of contextual information. These nine distinct views serve to address potential biases arising from multiple perspectives. Compared with Spectralformer and MFT, IF markedly improves performance. Even though IF, Spectralformer and MFT both employ the transformer as a backbone, Spectralformer and MFT do not fully explore the characteristics of different modalities, while IF can get comprehensive information for HSI multi-modal data sources.

\textbf{Evaluation on the MUUFL dataset} As shown in Table VI, OA of IF has yielded a favorable result than $4\%$ compared to the second-best IP-CNN. Similarly, the AA and the Kappa of IF perform the best at around $2\%$ and $5\%$ compared with IP-CNN, respectively, which means the IF framework has a remarkable ability to integrate different modalities from diverse perspectives coupled with contextual information. Compared to CNN-PPF, IF improves $7\%$, which means IF can combine the information of distinct modalities well without influencing the performance of a single HSI modality. Compared with Two-branch CNN, context CNN, Co-CNN, EndNet, MDL-RS, and S2FL, the IF framework gets competitive results because it considers the interaction across distinct modalities from diverse perspectives rather than traditional fusion from limited viewpoints or inadequate interactions of multi-modalities. Compared with Spectralformer and MFT, IF achieves competitive performance since IF contains a wealth of information between distinct modalities.

\textbf{Evaluation on the Trento dataset} As shown in Table VII, the result of IF framework is slightly lower than IP-CNN since IP-CNN needs two-stage training, including unsupervised and supervised training on the same dataset which can cause information leakage. Besides, IF may have an overfitting risk on the fusion matrix to pack plenty of interaction between different modalities. Compared with the other methods, the IF framework can get competitive results showing IF framework has the advantage of exploring comprehensive relationships for distinct modalities and eliminating bias from nine perspectives.   

Therefore, the IF framework can achieve the best performance in Trento, MUUFL, and Houston datasets without the risk of data leakage like IP-CNN. Moreover, IF gets competitive results compared to other fusion methods since IF extracts comprehensive characteristics from distinct modalities through nine interconnected perspectives in the fusion matrix. These nine interrelated elements can complement each other to remove the potential bias. In contrast, other fusion methods only employ traditional fusion methods from a limited viewpoint, which can cause insufficient information for different modalities and contain discrimination from limited views. Therefore, IF can create a novel insight for HSI multi-modal data fusion.

\begin{table*}[]
\centering
\caption{We compare IF with CNN-PPF, Two-branch
CNN, Context CNN, EndNet, MDL-RS, S2FL, Co-CNN, Spectralformer, MFT, and IP-CNN on the Houston dataset.}
\label{table6}
\setlength{\tabcolsep}{0.5mm}{
\begin{tabular}{cc|cccccccccccc}
\hline
\bottomrule
\multicolumn{1}{c|}{\multirow{2}{*}{}} & \multirow{2}{*}{} & \multicolumn{11}{c}{Performance}  \\ \cline{3-14} 

\multicolumn{1}{c|}{No.} &  Class                 & \multicolumn{1}{c|}{CNN-PPF\cite{li2016hyperspectral}} & \multicolumn{1}{c|}{ \makecell[c]{Two-branch  \\ CNN\cite{xu2017multisource}}}  & \multicolumn{1}{c|}{ \makecell[c]{Two-branch  \\ CNN (Merge)}
} & \multicolumn{1}{c|}{\makecell[c]{Context \\ CNN\cite{lee2017going}}} & \multicolumn{1}{c|}{EndNet\cite{hong2020deep}} & \multicolumn{1}{c|}{ \makecell[c]{MDL-\\RS\cite{hong2020more}}} & \multicolumn{1}{c|}{S2FL\cite{hong2021multimodal}} & \multicolumn{1}{c|}{\makecell[c]{Co-\\CNN \cite{hang2020classification}}} & \multicolumn{1}{c|}{\makecell[c]{Spectral\\former\cite{hong2021spectralformer}}}& \multicolumn{1}{c|}{MFT\cite{roy2022multimodal}}  & \multicolumn{1}{c|}{\makecell[c]{IP-\\CNN\cite{zhang2021information}}}  & IF  \\ \hline
\multicolumn{1}{c|}{1} & Health grass & \multicolumn{1}{c|}{83.57} & \multicolumn{1}{c|}{83.10}  & \multicolumn{1}{c|}{83.10} & \multicolumn{1}{c|}{84.89} & \multicolumn{1}{c|}{81.58} & \multicolumn{1}{c|}{83.10} & \multicolumn{1}{c|}{81.67} & \multicolumn{1}{c|}{\color{red}{98.51}} & \multicolumn{1}{c|}{82.65}  & \multicolumn{1}{c|}{82.34} & \multicolumn{1}{c|}{85.77} & \color{blue}{96.58} \\ \hline
\multicolumn{1}{c|}{2} & Stressed grass & \multicolumn{1}{c|}{98.21} & \multicolumn{1}{c|}{84.10} & \multicolumn{1}{c|}{81.20} & \multicolumn{1}{c|}{87.40} & \multicolumn{1}{c|}{83.65} & \multicolumn{1}{c|}{81.58} & \multicolumn{1}{c|}{82.99} & \multicolumn{1}{c|}{\color{blue}{97.83}} & \multicolumn{1}{c|}{83.33} & \multicolumn{1}{c|}{88.78}& \multicolumn{1}{c|}{87.34} & \color{red}{99.44} \\ \hline
\multicolumn{1}{c|}{3} & Synthetic grass & \multicolumn{1}{c|}{98.42} & \multicolumn{1}{c|}{\color{red}{100.00}} & \multicolumn{1}{c|}{\color{red}{100.00}} & \multicolumn{1}{c|}{\color{blue}{99.86}} & \multicolumn{1}{c|}{\color{red}{100.00}} & \multicolumn{1}{c|}{\color{red}{100.00}} & \multicolumn{1}{c|}{\color{red}{100.00}} & \multicolumn{1}{c|}{70.60} & \multicolumn{1}{c|}{75.78}& \multicolumn{1}{c|}{98.15} & \multicolumn{1}{c|}{\color{red}{100.00}} & 91.49 \\ \hline
\multicolumn{1}{c|}{4} & Tress  & \multicolumn{1}{c|}{\color{blue}{97.73}} & \multicolumn{1}{c|}{93.09}  & \multicolumn{1}{c|}{92.90} & \multicolumn{1}{c|}{93.49} & \multicolumn{1}{c|}{93.09} & \multicolumn{1}{c|}{\color{red}{99.72}} & \multicolumn{1}{c|}{92.05} & \multicolumn{1}{c|}{99.06} & \multicolumn{1}{c|}{91.10} & \multicolumn{1}{c|}{94.35} &\multicolumn{1}{c|}{94.26} & 95.64 \\ \hline
\multicolumn{1}{c|}{5} & Soil & \multicolumn{1}{c|}{96.50} & \multicolumn{1}{c|}{\color{red}{100.00}} & \multicolumn{1}{c|}{99.81} &  \multicolumn{1}{c|}{99.72} & \multicolumn{1}{c|}{\color{blue}{99.91}} & \multicolumn{1}{c|}{99.81} & \multicolumn{1}{c|}{99.43} & \multicolumn{1}{c|}{\color{red}{100.00}} & \multicolumn{1}{c|}{98.30} & \multicolumn{1}{c|}{99.12}& \multicolumn{1}{c|}{98.42} & 99.72 \\ \hline
\multicolumn{1}{c|}{6} & Water & \multicolumn{1}{c|}{97.20} & \multicolumn{1}{c|}{99.30} & \multicolumn{1}{c|}{\color{red}{100.00}} & \multicolumn{1}{c|}{98.77} & \multicolumn{1}{c|}{95.10} & \multicolumn{1}{c|}{95.10} & \multicolumn{1}{c|}{99.30} & \multicolumn{1}{c|}{41.11} & \multicolumn{1}{c|}{89.04} & \multicolumn{1}{c|}{99.30} & \multicolumn{1}{c|}{\color{blue}{99.91}} & 92.31 \\ \hline
\multicolumn{1}{c|}{7} & Residential & \multicolumn{1}{c|}{85.82} & \multicolumn{1}{c|}{92.82} & \multicolumn{1}{c|}{92.54} & \multicolumn{1}{c|}{82.81} & \multicolumn{1}{c|}{82.65} & \multicolumn{1}{c|}{90.02} & \multicolumn{1}{c|}{76.31} & \multicolumn{1}{c|}{83.14} & \multicolumn{1}{c|}{81.72} & \multicolumn{1}{c|}{88.56} & \multicolumn{1}{c|}{\color{blue}{94.59}} & \color{red}{98.23} \\ \hline
\multicolumn{1}{c|}{8} & Commercial & \multicolumn{1}{c|}{56.51} & \multicolumn{1}{c|}{82.34} & \multicolumn{1}{c|}{\color{blue}{94.87}} & \multicolumn{1}{c|}{78.78} & \multicolumn{1}{c|}{81.29} & \multicolumn{1}{c|}{87.94} & \multicolumn{1}{c|}{68.09} & \multicolumn{1}{c|}{98.39} & \multicolumn{1}{c|}{67.81}  & \multicolumn{1}{c|}{86.89} &\multicolumn{1}{c|}{91.81} & \color{red}{99.43} \\ \hline
\multicolumn{1}{c|}{9} & Road & \multicolumn{1}{c|}{71.20} & \multicolumn{1}{c|}{84.70}  & \multicolumn{1}{c|}{83.85} & \multicolumn{1}{c|}{82.51} & \multicolumn{1}{c|}{88.29} & \multicolumn{1}{c|}{81.59} & \multicolumn{1}{c|}{69.22} & \multicolumn{1}{c|}{\color{blue}{94.81}} & \multicolumn{1}{c|}{74.47} & \multicolumn{1}{c|}{87.91} & \multicolumn{1}{c|}{89.35} &  \color{red}{95.18} \\ \hline
\multicolumn{1}{c|}{10} & Highway & \multicolumn{1}{c|}{57.12} & \multicolumn{1}{c|}{65.44}  & \multicolumn{1}{c|}{69.89} & \multicolumn{1}{c|}{59.41} & \multicolumn{1}{c|}{89.00} & \multicolumn{1}{c|}{86.68} & \multicolumn{1}{c|}{52.12} & \multicolumn{1}{c|}{\color{blue}{92.98}} & \multicolumn{1}{c|}{56.76}& \multicolumn{1}{c|}{64.70} & \multicolumn{1}{c|}{72.43} & \color{red}{98.65} \\ \hline
\multicolumn{1}{c|}{11} & Railway & \multicolumn{1}{c|}{80.55} & \multicolumn{1}{c|}{88.24}  & \multicolumn{1}{c|}{86.15} & \multicolumn{1}{c|}{83.24} & \multicolumn{1}{c|}{83.78} & \multicolumn{1}{c|}{89.37} & \multicolumn{1}{c|}{90.42} & \multicolumn{1}{c|}{90.88} & \multicolumn{1}{c|}{59.93} & \multicolumn{1}{c|}{\color{blue}{98.64}} & \multicolumn{1}{c|}{96.57} & \color{red}{99.62} \\ \hline
\multicolumn{1}{c|}{12} & Parking lot 1 & \multicolumn{1}{c|}{62.82} & \multicolumn{1}{c|}{89.53}  & \multicolumn{1}{c|}{92.60} & \multicolumn{1}{c|}{92.13} & \multicolumn{1}{c|}{90.39} & \multicolumn{1}{c|}{85.69} & \multicolumn{1}{c|}{87.42} & \multicolumn{1}{c|}{91.02} & \multicolumn{1}{c|}{70.00} & \multicolumn{1}{c|}{94.24} & \multicolumn{1}{c|}{\color{blue}{95.60}} & \color{red}{99.23} \\ \hline
\multicolumn{1}{c|}{13} & Parking lot 2 & \multicolumn{1}{c|}{63.86} & \multicolumn{1}{c|}{92.28} & \multicolumn{1}{c|}{79.30} & \multicolumn{1}{c|}{\color{red}{94.88}} & \multicolumn{1}{c|}{82.46} & \multicolumn{1}{c|}{83.16} & \multicolumn{1}{c|}{70.18} & \multicolumn{1}{c|}{97.09} & \multicolumn{1}{c|}{66.20} & \multicolumn{1}{c|}{90.29} & \multicolumn{1}{c|}{94.37} & \color{blue}{94.39} \\ \hline
\multicolumn{1}{c|}{14} & Tennis court & \multicolumn{1}{c|}{\color{red}{100.00}} & \multicolumn{1}{c|}{96.76}  & \multicolumn{1}{c|}{\color{red}{100.00}} & \multicolumn{1}{c|}{99.77} & \multicolumn{1}{c|}{\color{red}{100.00}} & \multicolumn{1}{c|}{\color{red}{100.00}} & \multicolumn{1}{c|}{\color{red}{100.00}} & \multicolumn{1}{c|}{\color{red}{100.00}} & \multicolumn{1}{c|}{92.04}& \multicolumn{1}{c|}{99.73} & \multicolumn{1}{c|}{\color{blue}{99.86}} & 99.60 \\ \hline

\multicolumn{1}{c|}{15} & Running track & \multicolumn{1}{c|}{98.10} & \multicolumn{1}{c|}{99.79}  & \multicolumn{1}{c|}{\color{red}{100.00}} & \multicolumn{1}{c|}{98.79} & \multicolumn{1}{c|}{98.10} & \multicolumn{1}{c|}{98.73} & \multicolumn{1}{c|}{98.31} & \multicolumn{1}{c|}{97.85} & \multicolumn{1}{c|}{77.45}  & \multicolumn{1}{c|}{99.58} & \multicolumn{1}{c|}{\color{blue}{99.99}} & 91.33 \\ \hline
\multicolumn{2}{c|}{OA (\%)}  & \multicolumn{1}{c|}{83.33} & \multicolumn{1}{c|}{87.98}  & \multicolumn{1}{c|}{88.91} & \multicolumn{1}{c|}{86.90} & \multicolumn{1}{c|}{88.52} & \multicolumn{1}{c|}{89.60} & \multicolumn{1}{c|}{81.95} & \multicolumn{1}{c|}{90.43} & \multicolumn{1}{c|}{76.87}& \multicolumn{1}{c|}{89.80} & \multicolumn{1}{c|}{\color{blue}{92.06}} & \color{red}{97.50} \\ \hline
\multicolumn{2}{c|}{AA (\%)} & \multicolumn{1}{c|}{83.21} & \multicolumn{1}{c|}{90.11} & \multicolumn{1}{c|}{90.42} & \multicolumn{1}{c|}{89.11} & \multicolumn{1}{c|}{89.95} & \multicolumn{1}{c|}{90.83} & \multicolumn{1}{c|}{84.50} & \multicolumn{1}{c|}{90.22} & \multicolumn{1}{c|}{77.77} & \multicolumn{1}{c|}{91.51} & \multicolumn{1}{c|}{\color{blue}{93.35}} & \color{red}{96.72} \\ \hline
\multicolumn{2}{c|}{Kappa (\%)} & \multicolumn{1}{c|}{81.88} & \multicolumn{1}{c|}{86.98}  & \multicolumn{1}{c|}{87.96} & \multicolumn{1}{c|}{85.89} & \multicolumn{1}{c|}{87.59} & \multicolumn{1}{c|}{88.75} & \multicolumn{1}{c|}{80.40} & \multicolumn{1}{c|}{89.68} & \multicolumn{1}{c|}{75.03}& \multicolumn{1}{c|}{88.93} & \multicolumn{1}{c|}{\color{blue}{91.42}} & \color{red}{97.29}\\ \hline
\bottomrule
\end{tabular}}
\end{table*}

\begin{table*}[]
\centering
\caption{We compare IF with CNN-PPF, Two-branch
CNN, Context CNN, EndNet, MDL-RS, S2FL, Co-CNN, Spectralformer, MFT, and IP-CNN on the MUUFL dataset.}
\label{table4}
\setlength{\tabcolsep}{0.5mm}{
\begin{tabular}{cc|cccccccccccc}
\hline
\bottomrule
\multicolumn{1}{c|}{\multirow{2}{*}{}} & \multirow{2}{*}{} & \multicolumn{11}{c}{Performance}  \\ \cline{3-14} 
\multicolumn{1}{c|}{No.} &  Class                 & \multicolumn{1}{c|}{ \makecell[c]{CNN\\-PPF\cite{li2016hyperspectral}}} & \multicolumn{1}{c|}{ \makecell[c]{Two-branch  \\ CNN\cite{xu2017multisource}}} & \multicolumn{1}{c|}{ \makecell[c]{Two-branch  \\ CNN (Merge)}
} & \multicolumn{1}{c|}{ \makecell[c]{Context \\ CNN\cite{lee2017going}}} & \multicolumn{1}{c|}{ \makecell[c]{EndNet\\\cite{hong2020deep}}} & \multicolumn{1}{c|}{ \makecell[c]{MDL-\\RS\cite{hong2020more}}} & \multicolumn{1}{c|}{ \makecell[c]{S2FL\\\cite{hong2021multimodal}}} & \multicolumn{1}{c|}{\makecell[c]{ \makecell[c]{Co-\\CNN\cite{hang2020classification}}}} & \multicolumn{1}{c|}{ \makecell[c]{Spectral\\former\cite{hong2021spectralformer}}}& \multicolumn{1}{c|}{MFT\cite{roy2022multimodal}}  & \multicolumn{1}{c|}{ \makecell[c]{IP-\\CNN\cite{zhang2021information}}} & IF  \\ \hline
\multicolumn{1}{c|}{1}  &   Trees                & \multicolumn{1}{c|}{89.07} & \multicolumn{1}{c|}{92.35}  & \multicolumn{1}{c|}{94.28} & \multicolumn{1}{c|}{91.29} & \multicolumn{1}{c|}{89.00} & \multicolumn{1}{c|}{88.57} & \multicolumn{1}{c|}{72.40} & \multicolumn{1}{c|}{\color{red}{98.90}} & \multicolumn{1}{c|}{97.30} & \multicolumn{1}{c|}{97.90} & \multicolumn{1}{c|}{94.40} & \color{blue}{98.87} \\ \hline
\multicolumn{1}{c|}{2}  &  Mostly grass       & \multicolumn{1}{c|}{85.71} & \multicolumn{1}{c|}{59.30}  & \multicolumn{1}{c|}{82.84} & \multicolumn{1}{c|}{63.09} & \multicolumn{1}{c|}{84.73} & \multicolumn{1}{c|}{84.00} & \multicolumn{1}{c|}{75.97} & \multicolumn{1}{c|}{78.60} & \multicolumn{1}{c|}{69.35} & \multicolumn{1}{c|}{92.11} & \multicolumn{1}{c|}{\color{blue}{92.26}} & \color{red}{96.12}  \\ \hline
\multicolumn{1}{c|}{3}  &  Mixed ground surface & \multicolumn{1}{c|}{80.15} & \multicolumn{1}{c|}{94.47}  & \multicolumn{1}{c|}{84.58} & \multicolumn{1}{c|}{81.84} & \multicolumn{1}{c|}{73.39} & \multicolumn{1}{c|}{73.79} & \multicolumn{1}{c|}{54.72} & \multicolumn{1}{c|}{90.66} & \multicolumn{1}{c|}{78.48}& \multicolumn{1}{c|}{\color{blue}{91.80}} & \multicolumn{1}{c|}{87.96} & \color{red}{95.94} \\ \hline
\multicolumn{1}{c|}{4}  &  Dirt and sand & \multicolumn{1}{c|}{93.10} & \multicolumn{1}{c|}{93.74}  & \multicolumn{1}{c|}{\color{blue}{97.32}} & \multicolumn{1}{c|}{93.92} & \multicolumn{1}{c|}{91.95} & \multicolumn{1}{c|}{93.48} & \multicolumn{1}{c|}{82.20} & \multicolumn{1}{c|}{90.60} & \multicolumn{1}{c|}{82.63} & \multicolumn{1}{c|}{91.59} & \multicolumn{1}{c|}{97.15} & \color{red}{99.16} \\ \hline
\multicolumn{1}{c|}{5} & Road & \multicolumn{1}{c|}{88.98} & \multicolumn{1}{c|}{92.76}  & \multicolumn{1}{c|}{92.96} & \multicolumn{1}{c|}{89.44} & \multicolumn{1}{c|}{87.72} & \multicolumn{1}{c|}{88.54} & \multicolumn{1}{c|}{71.26} & \multicolumn{1}{c|}{\color{blue}{96.90}} & \multicolumn{1}{c|}{87.91} & \multicolumn{1}{c|}{95.60} & \multicolumn{1}{c|}{94.38} & \color{red}{97.87} \\ \hline
\multicolumn{1}{c|}{6}  &   Water   & \multicolumn{1}{c|}{98.93} & \multicolumn{1}{c|}{98.42}  & \multicolumn{1}{c|}{99.05} & \multicolumn{1}{c|}{92.92} & \multicolumn{1}{c|}{\color{blue}{99.14}} & \multicolumn{1}{c|}{98.42} & \multicolumn{1}{c|}{94.42} & \multicolumn{1}{c|}{75.98} & \multicolumn{1}{c|}{58.77} & \multicolumn{1}{c|}{88.19} & \multicolumn{1}{c|}{\color{red}{99.79}} & 98.42 \\ \hline
\multicolumn{1}{c|}{7}  & Building shadow & \multicolumn{1}{c|}{89.07} & \multicolumn{1}{c|}{95.68} & \multicolumn{1}{c|}{93.28} & \multicolumn{1}{c|}{84.73} & \multicolumn{1}{c|}{91.58} & \multicolumn{1}{c|}{90.46} & \multicolumn{1}{c|}{77.34} & \multicolumn{1}{c|}{73.54} & \multicolumn{1}{c|}{85.87} & \multicolumn{1}{c|}{90.27} & \multicolumn{1}{c|}{\color{blue}{96.30}} & \color{red}{97.17} \\ \hline
\multicolumn{1}{c|}{8}  & Building & \multicolumn{1}{c|}{92.15} & \multicolumn{1}{c|}{94.01}  & \multicolumn{1}{c|}{95.42} & \multicolumn{1}{c|}{81.22} & \multicolumn{1}{c|}{91.30} & \multicolumn{1}{c|}{92.95} & \multicolumn{1}{c|}{86.19} & \multicolumn{1}{c|}{96.66} & \multicolumn{1}{c|}{95.60}& \multicolumn{1}{c|}{\color{blue}{97.26}} & \multicolumn{1}{c|}{96.13} & \color{red}{98.87} \\ \hline
\multicolumn{1}{c|}{9}  & Sidewalk & \multicolumn{1}{c|}{75.45} & \multicolumn{1}{c|}{86.64}  & \multicolumn{1}{c|}{84.13} & \multicolumn{1}{c|}{81.30} & \multicolumn{1}{c|}{83.75} & \multicolumn{1}{c|}{81.37} & \multicolumn{1}{c|}{59.21} & \multicolumn{1}{c|}{64.93} & \multicolumn{1}{c|}{53.52}& \multicolumn{1}{c|}{61.35} & \multicolumn{1}{c|}{\color{red}{94.01}} & \color{blue}{93.36} \\ \hline
\multicolumn{1}{c|}{10}  & Yellow curb & \multicolumn{1}{c|}{\color{red}{100.00}} & \multicolumn{1}{c|}{\color{red}{100.00}}  & \multicolumn{1}{c|}{\color{red}{100.00}} & \multicolumn{1}{c|}{98.91} & \multicolumn{1}{c|}{\color{red}{100.00}} & \multicolumn{1}{c|}{\color{red}{100.00}} & \multicolumn{1}{c|}{\color{blue}{98.91}} & \multicolumn{1}{c|}{19.47} & \multicolumn{1}{c|}{08.43} & \multicolumn{1}{c|}{17.43} &  \multicolumn{1}{c|}{\color{red}{100.00}} & 96.97 \\ \hline
\multicolumn{1}{c|}{11} & Cloth panels & \multicolumn{1}{c|}{\color{red}{100.00}} & \multicolumn{1}{c|}{96.64}  & \multicolumn{1}{c|}{97.48} & \multicolumn{1}{c|}{99.63} & \multicolumn{1}{c|}{99.63} & \multicolumn{1}{c|}{99.26} & \multicolumn{1}{c|}{98.88} & \multicolumn{1}{c|}{62.76} & \multicolumn{1}{c|}{35.29}& \multicolumn{1}{c|}{72.79} & \multicolumn{1}{c|}{\color{blue}{99.63}} & 99.16 \\ \hline
\multicolumn{2}{c|}{OA (\%)}  & \multicolumn{1}{c|}{90.97} & \multicolumn{1}{c|}{90.35}  & \multicolumn{1}{c|}{91.95} & \multicolumn{1}{c|}{86.07} & \multicolumn{1}{c|}{87.02} & \multicolumn{1}{c|}{87.54} & \multicolumn{1}{c|}{72.49} & \multicolumn{1}{c|}{90.93} & \multicolumn{1}{c|}{88.25} & \multicolumn{1}{c|}{\color{blue}{94.34}} & \multicolumn{1}{c|}{93.86} & \color{red}{97.96} \\ \hline
\multicolumn{2}{c|}{AA (\%)}  & \multicolumn{1}{c|}{90.24} & \multicolumn{1}{c|}{91.27}  & \multicolumn{1}{c|}{92.58} & \multicolumn{1}{c|}{87.12} & \multicolumn{1}{c|}{90.20} & \multicolumn{1}{c|}{90.14} & \multicolumn{1}{c|}{79.23} & \multicolumn{1}{c|}{77.18} & \multicolumn{1}{c|}{68.47}& \multicolumn{1}{c|}{81.48} & \multicolumn{1}{c|}{\color{blue}{95.64}} & \color{red}{97.45} \\ \hline
\multicolumn{2}{c|}{Kappa (\%)} & \multicolumn{1}{c|}{84.46} & \multicolumn{1}{c|}{87.27}  & \multicolumn{1}{c|}{89.35} & \multicolumn{1}{c|}{81.89} & \multicolumn{1}{c|}{83.24} & \multicolumn{1}{c|}{83.31} & \multicolumn{1}{c|}{65.81} & \multicolumn{1}{c|}{88.22} & \multicolumn{1}{c|}{84.40}& \multicolumn{1}{c|}{\color{blue}{92.51}} & \multicolumn{1}{c|}{91.99} & \color{red}{97.27} \\ \hline
\bottomrule
\end{tabular}}
\end{table*}


\begin{table*}[]
\centering
\caption{We compare IF with CNN-PPF, Two-branch
CNN, Context CNN, EndNet, MDL-RS, S2FL, Co-CNN, Spectralformer, MFT, and IP-CNN on Trento dataset. More specifically, IP-CNN needs two-stage training including unsupervised and supervised training which will cause information leakage. }
\label{table4}
\setlength{\tabcolsep}{0.75mm}{
\begin{tabular}{cc|cccccccccccc}
\hline
\bottomrule
\multicolumn{1}{c|}{\multirow{2}{*}{}} & \multirow{2}{*}{} & \multicolumn{11}{c}{Performance}  \\ \cline{3-14} 
\multicolumn{1}{c|}{No.} &  Class                 & \multicolumn{1}{c|}{ \makecell[c]{CNN\\-PPF\cite{li2016hyperspectral}}} & \multicolumn{1}{c|}{ \makecell[c]{Two-branch  \\ CNN\cite{xu2017multisource}}} & \multicolumn{1}{c|}{ \makecell[c]{Two-branch  \\ CNN (Merge)}
} & \multicolumn{1}{c|}{ \makecell[c]{Context \\ CNN\cite{lee2017going}}} & \multicolumn{1}{c|}{\makecell[c]{EndNet\\\cite{hong2020deep}}} & \multicolumn{1}{c|}{ \makecell[c]{MDL-\\RS\cite{hong2020more}}} & \multicolumn{1}{c|}{\makecell[c]{S2FL\\\cite{hong2021multimodal}}} & \multicolumn{1}{c|}{ \makecell[c]{Co-\\CNN\cite{hang2020classification}}} & \multicolumn{1}{c|}{ \makecell[c]{Spectral\\former\cite{hong2021spectralformer}}}& \multicolumn{1}{c|}{MFT\cite{roy2022multimodal}}  & \multicolumn{1}{c|}{ \makecell[c]{IP-\\CNN\cite{zhang2021information}}} & IF  \\ \hline
\multicolumn{1}{c|}{1} & Apple trees & \multicolumn{1}{c|}{90.11} & \multicolumn{1}{c|}{98.07} & \multicolumn{1}{c|}{98.51} & \multicolumn{1}{c|}{\color{blue}{99.26}} & \multicolumn{1}{c|}{88.19} & \multicolumn{1}{c|}{88.58} & \multicolumn{1}{c|}{79.26} & \multicolumn{1}{c|}{\color{red}{99.87}} & \multicolumn{1}{c|}{96.76} & \multicolumn{1}{c|}{98.23}& \multicolumn{1}{c|}{99.00} & 99.15 \\ \hline
\multicolumn{1}{c|}{2} & Buildings & \multicolumn{1}{c|}{83.34} & \multicolumn{1}{c|}{95.21} & \multicolumn{1}{c|}{92.49} & \multicolumn{1}{c|}{86.81} & \multicolumn{1}{c|}{98.49} & \multicolumn{1}{c|}{95.86} & \multicolumn{1}{c|}{92.44} & \multicolumn{1}{c|}{83.84} & \multicolumn{1}{c|}{97.25}& \multicolumn{1}{c|}{\color{blue}{99.34}} & \multicolumn{1}{c|}{\color{red}{99.40}} & 98.92 \\ \hline
\multicolumn{1}{c|}{3} & Ground & \multicolumn{1}{c|}{71.13} & \multicolumn{1}{c|}{93.32}  & \multicolumn{1}{c|}{\color{red}{100.00}} & \multicolumn{1}{c|}{97.91} & \multicolumn{1}{c|}{95.19} & \multicolumn{1}{c|}{93.58} & \multicolumn{1}{c|}{92.78} & \multicolumn{1}{c|}{87.09} & \multicolumn{1}{c|}{58.47}& \multicolumn{1}{c|}{89.84} & \multicolumn{1}{c|}{\color{blue}{99.10}} & 84.49 \\ \hline
\multicolumn{1}{c|}{4} & Woods & \multicolumn{1}{c|}{99.04} & \multicolumn{1}{c|}{99.93}  & \multicolumn{1}{c|}{97.32} & \multicolumn{1}{c|}{97.31} & \multicolumn{1}{c|}{99.30} & \multicolumn{1}{c|}{99.22} & \multicolumn{1}{c|}{94.94} & \multicolumn{1}{c|}{\color{blue}{99.98}} & \multicolumn{1}{c|}{99.24}& \multicolumn{1}{c|}{99.82} & \multicolumn{1}{c|}{99.92} & \color{red}{100.00} \\ \hline
\multicolumn{1}{c|}{5} & Vineyard & \multicolumn{1}{c|}{99.37} & \multicolumn{1}{c|}{98.78}  & \multicolumn{1}{c|}{\color{red}{100.00}} & \multicolumn{1}{c|}{99.82} & \multicolumn{1}{c|}{91.96} & \multicolumn{1}{c|}{83.82} & \multicolumn{1}{c|}{72.83} & \multicolumn{1}{c|}{99.61} & \multicolumn{1}{c|}{93.52} & \multicolumn{1}{c|}{99.93} & \multicolumn{1}{c|}{99.66} & \color{blue}{99.94} \\ \hline
\multicolumn{1}{c|}{6} & Roads & \multicolumn{1}{c|}{89.73} & \multicolumn{1}{c|}{89.98}  & \multicolumn{1}{c|}{\color{blue}{92.56}} & \multicolumn{1}{c|}{84.63} & \multicolumn{1}{c|}{90.14} & \multicolumn{1}{c|}{76.51} & \multicolumn{1}{c|}{86.40} & \multicolumn{1}{c|}{\color{red}{98.75}} & \multicolumn{1}{c|}{73.39} & \multicolumn{1}{c|}{88.72} & \multicolumn{1}{c|}{90.21} & 90.14 \\ \hline
\multicolumn{2}{c|}{OA (\%)} & \multicolumn{1}{c|}{94.76} & \multicolumn{1}{c|}{97.92}  & \multicolumn{1}{c|}{97.46} & \multicolumn{1}{c|}{96.11} & \multicolumn{1}{c|}{94.17} & \multicolumn{1}{c|}{90.65} & \multicolumn{1}{c|}{83.95} & \multicolumn{1}{c|}{97.69} & \multicolumn{1}{c|}{93.51}& \multicolumn{1}{c|}{98.32} & \multicolumn{1}{c|}{\color{red}{98.58}} & \color{blue}{98.54} \\ \hline
\multicolumn{2}{c|}{AA (\%)} & \multicolumn{1}{c|}{88.97} & \multicolumn{1}{c|}{96.19}  & \multicolumn{1}{c|}{96.80} & \multicolumn{1}{c|}{94.29} & \multicolumn{1}{c|}{93.88} & \multicolumn{1}{c|}{89.60} & \multicolumn{1}{c|}{86.44} & \multicolumn{1}{c|}{94.86} & \multicolumn{1}{c|}{86.44}& \multicolumn{1}{c|}{\color{blue}{95.98}} & \multicolumn{1}{c|}{\color{red}{97.88}} & 95.44 \\ \hline
\multicolumn{2}{c|}{Kappa (\%)} & \multicolumn{1}{c|}{93.04} & \multicolumn{1}{c|}{96.81} & \multicolumn{1}{c|}{96.61} & \multicolumn{1}{c|}{94.81} & \multicolumn{1}{c|}{92.22} & \multicolumn{1}{c|}{86.28} & \multicolumn{1}{c|}{78.99} & \multicolumn{1}{c|}{96.91} & \multicolumn{1}{c|}{91.36}& \multicolumn{1}{c|}{97.75} & \multicolumn{1}{c|}{\color{red}{98.17}} & \color{blue}{98.05}  \\ \hline
\bottomrule
\end{tabular}}
\end{table*}

\subsection{Experiments with fusion strategies and patch sizes}
In addition, we explore the impact of the change in
patch sizes and fusion strategies on the final performance. We compare the performance of four distinct size settings of input data sources, including $3\times 3$, $6\times 6$, $9\times 9$, and $12\times 12$, with three different fusion strategies settings, including early, middle, and late. In stage 1, we set the number of patches for the input as nine in all situations, which means the small patch size for $3\times 3$ is on the pixel level, while $6\times 6$, $9\times 9$ and $12\times 12$ are on patch level. As shown in Tables VIII, IX, and X, the $3\times 3$ achieved the worst performance among different fusion strategies in three datasets since pixel-level input has insufficient neighborhood information to explore the relationship between the center patch and nearby patches for intra- and inter-modalities. When patch size increases, there is a clear increasing tendency, which means the patch size positively affects final performance. It is worth noting that the IF framework can achieve similar performance regardless of distinct fusion strategies when input is on the patch level, which means the IF framework can achieve robust performance on different fusion strategies.


\begin{table*}[]
\centering
\caption{The Houston dataset performance change follows the change of different fusion stages and the size of input image }
\label{table8}
\begin{tabular}{c|cccc|cccc|cccc}
\hline
\bottomrule
 &                      &     Early                   &                       &  &                       &   Middle                   &                       &  &                       & Late                      &                       &  \\ \cline{2-13} 
 Metrix & \multicolumn{1}{c|}{$3\times 3$} & \multicolumn{1}{c|}{$6\times 6$} & \multicolumn{1}{c|}{$9\times 9$} & $12\times 12$  & \multicolumn{1}{c|}{$3\times 3$} & \multicolumn{1}{c|}{$6\times 6$} & \multicolumn{1}{c|}{$9\times 9$} & $12\times 12$  & \multicolumn{1}{c|}{$3\times 3$} & \multicolumn{1}{c|}{$6\times 6$} & \multicolumn{1}{c|}{$9\times 9$} &$12\times 12$  \\ \hline
 OA & \multicolumn{1}{c|}{85.67} & \multicolumn{1}{c|}{94.36} & \multicolumn{1}{c|}{96.37} & 97.50  & \multicolumn{1}{c|}{84.82} & \multicolumn{1}{c|}{94.02} & \multicolumn{1}{c|}{95.95} & 96.38  & \multicolumn{1}{c|}{82.74} & \multicolumn{1}{c|}{94.78} & \multicolumn{1}{c|}{93.69} & 97.03  \\ \hline
 AA & \multicolumn{1}{c|}{86.05} & \multicolumn{1}{c|}{93.89} & \multicolumn{1}{c|}{95.42} & 96.72  & \multicolumn{1}{c|}{84.87} & \multicolumn{1}{c|}{93.25} & \multicolumn{1}{c|}{95.22} & 95.83 & \multicolumn{1}{c|}{83.11} & \multicolumn{1}{c|}{94.46} & \multicolumn{1}{c|}{93.15} & 96.52 \\ \hline
Kappa & \multicolumn{1}{c|}{84.46} & \multicolumn{1}{c|}{93.88} & \multicolumn{1}{c|}{96.06} & 97.29 & \multicolumn{1}{c|}{83.52} & \multicolumn{1}{c|}{93.51} & \multicolumn{1}{c|}{95.60} & 96.07 & \multicolumn{1}{c|}{81.27} & \multicolumn{1}{c|}{94.33} & \multicolumn{1}{c|}{93.15} & 96.78 \\ \hline
\end{tabular}
\end{table*}

\begin{table*}[]
\centering
\caption{The Trento dataset performance change follows the change of different fusion stages and the size of input image}
\label{table8}
\begin{tabular}{c|cccc|cccc|cccc}
\hline
\bottomrule
 &                      &     Early                   &                       &  &                       &   Middle                   &                       &  &                       & Late                      &                       &  \\ \cline{2-13} 
Metrix & \multicolumn{1}{c|}{$3\times 3$} & \multicolumn{1}{c|}{$6\times 6$} & \multicolumn{1}{c|}{$9\times 9$} & $12\times 12$  & \multicolumn{1}{c|}{$3\times 3$} & \multicolumn{1}{c|}{$6\times 6$} & \multicolumn{1}{c|}{$9\times 9$} & $12\times 12$  & \multicolumn{1}{c|}{$3\times 3$} & \multicolumn{1}{c|}{$6\times 6$} & \multicolumn{1}{c|}{$9\times 9$} &$12\times 12$  \\ \hline
 OA & \multicolumn{1}{c|}{97.28} & \multicolumn{1}{c|}{98.24} & \multicolumn{1}{c|}{98.54} & 98.39  & \multicolumn{1}{c|}{97.05} & \multicolumn{1}{c|}{97.51} & \multicolumn{1}{c|}{98.50} & 98.33  & \multicolumn{1}{c|}{97.20} & \multicolumn{1}{c|}{97.81} & \multicolumn{1}{c|}{98.37} & 98.32 \\ \hline
 AA & \multicolumn{1}{c|}{93.50} & \multicolumn{1}{c|}{96.18} & \multicolumn{1}{c|}{95.44} & 96.41  & \multicolumn{1}{c|}{93.18} & \multicolumn{1}{c|}{92.91} & \multicolumn{1}{c|}{95.97} & 96.00  & \multicolumn{1}{c|}{93.00} & \multicolumn{1}{c|}{93.82} & \multicolumn{1}{c|}{96.04} & 95.45 \\ \hline
Kappa & \multicolumn{1}{c|}{96.35} & \multicolumn{1}{c|}{97.65} & \multicolumn{1}{c|}{98.05} & 97.85  & \multicolumn{1}{c|}{96.04} & \multicolumn{1}{c|}{96.66} & \multicolumn{1}{c|}{97.99} & 97.76  & \multicolumn{1}{c|}{96.24} & \multicolumn{1}{c|}{97.07} & \multicolumn{1}{c|}{97.81} & 97.75 \\ \hline
\end{tabular}
\end{table*}

\begin{table*}[]
\centering
\caption{The Muffl dataset performance change follows the change of different fusion stages and the size of input image}
\label{table8}
\begin{tabular}{c|cccc|cccc|cccc}
\hline
\bottomrule
 &                      &     Early                   &                       &  &                       &   Middle                   &                       &  &                       & Late                      &                       &  \\ \cline{2-13} 
 Metrix & \multicolumn{1}{c|}{$3\times 3$} & \multicolumn{1}{c|}{$6\times 6$} & \multicolumn{1}{c|}{$9\times 9$} & $12\times 12$  & \multicolumn{1}{c|}{$3\times 3$} & \multicolumn{1}{c|}{$6\times 6$} & \multicolumn{1}{c|}{$9\times 9$} & $12\times 12$  & \multicolumn{1}{c|}{$3\times 3$} & \multicolumn{1}{c|}{$6\times 6$} & \multicolumn{1}{c|}{$9\times 9$} &$12\times 12$  \\ \hline
 OA & \multicolumn{1}{c|}{95.49} & \multicolumn{1}{c|}{97.08} & \multicolumn{1}{c|}{97.48} & 97.84  & \multicolumn{1}{c|}{95.89} & \multicolumn{1}{c|}{97.29} & \multicolumn{1}{c|}{97.73} & 97.96 & \multicolumn{1}{c|}{95.74} & \multicolumn{1}{c|}{97.09} & \multicolumn{1}{c|}{97.48} & 97.78 \\ \hline
 AA & \multicolumn{1}{c|}{95.56} & \multicolumn{1}{c|}{96.55} & \multicolumn{1}{c|}{97.14} & 97.54  & \multicolumn{1}{c|}{95.42} & \multicolumn{1}{c|}{96.79} & \multicolumn{1}{c|}{97.51} & 97.45  & \multicolumn{1}{c|}{95.05} & \multicolumn{1}{c|}{97.04} & \multicolumn{1}{c|}{97.14} & 97.51 \\ \hline
Kappa & \multicolumn{1}{c|}{94.00} & \multicolumn{1}{c|}{96.11} & \multicolumn{1}{c|}{96.63} & 97.11 & \multicolumn{1}{c|}{94.50} & \multicolumn{1}{c|}{96.38} & \multicolumn{1}{c|}{96.97} & 97.27 & \multicolumn{1}{c|}{94.31} & \multicolumn{1}{c|}{96.11} & \multicolumn{1}{c|}{96.63} & 97.04 \\ \hline
\end{tabular}
\end{table*}

\subsection{Ablation studies}

\begin{figure*}[!ht]
  \centering
  \subfigure[\normalsize Water]{
  \includegraphics[width=1.2\columnwidth]{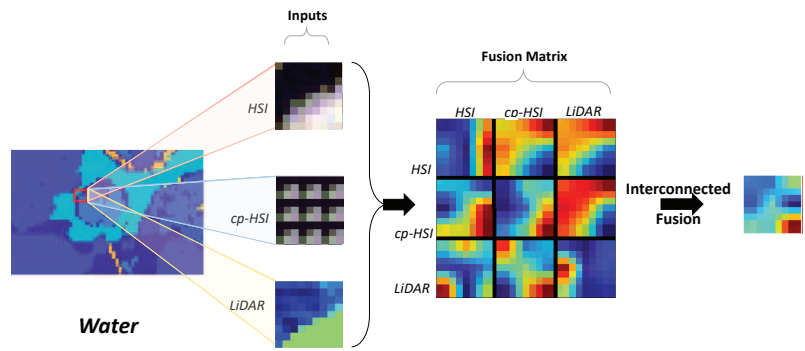}
  }
  \subfigure[\normalsize Soil]{
  \includegraphics[width=1.2\columnwidth]{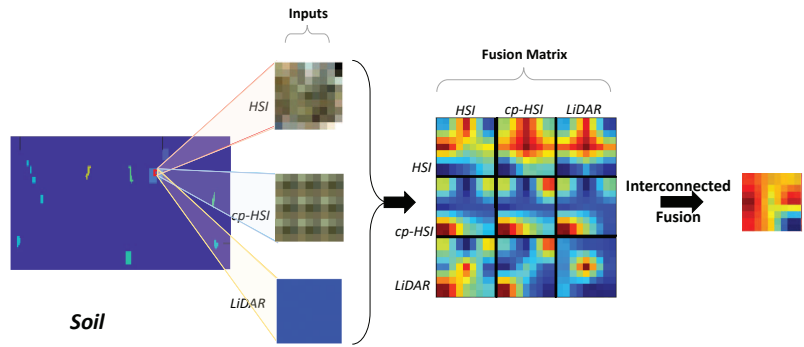}
  }
  \subfigure[\normalsize Sidewalk]{
  \includegraphics[width=1.2\columnwidth]{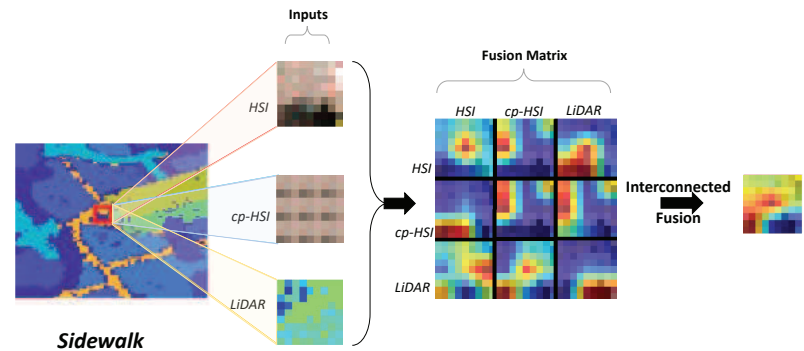}
  }
  \subfigure[\normalsize Wood]{
  \includegraphics[width=1.2\columnwidth]{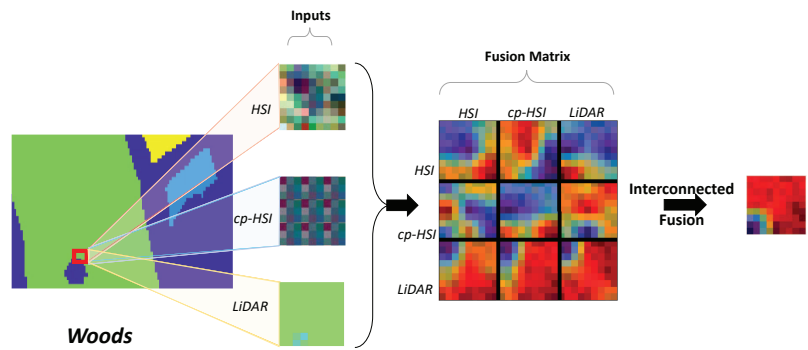}
  }
  \vspace{1pt}
  \centering
  \caption{\normalsize The three inputs, fusion matrix, and integrated feature after the Interconnected Fusion stage have been visualized.
 The first input shows the false color visualization of the HSI input. The second input is the center patch of HSI. The third input represents the corresponding LiDAR visualization (blue represents no altitude information). Then the nine perspectives in the fusion matrix have been calculated and visualized. After the Interconnected Fusion Stage, the nine interconnected elements can complement each other and eliminate bias from each perspective.}
  \label{fig:figure}
\end{figure*}

\begin{table*}[]
\centering
\caption{The classification performance of different ablation settings is conducted here. Without the contextual information means removing the center-patch intra-modality feature encoding module in Stage 1. Without fusion matrix indicates combining the nine elements in the fusion matrix through concatenation. In addition, HSI modality represents implementing a model merely based on HSI.}
\label{table8}
\begin{tabular}{c|c|c|c|c|c}
\hline
\bottomrule
                  & Metrix   &  \makecell[c]{Without contextual \\ information}  & \makecell[c]{Without \\ fusion matrix }  &  \makecell[c]{HSI modality} & IF framework  \\ \hline
\multirow{3}{*}{Houston} & OA  & 96.41  & 91.35  & 90.31 & \textbf{97.50}\\ \cline{2-6}
                  & AA & 95.76   & 95.31  & 90.10  & \textbf{96.72}\\ \cline{2-6} 
                  & Kappa & 96.10   & 95.56  & 88.96 & \textbf{97.29} \\ \hline
\multirow{3}{*}{Trento} & OA    & 97.73  & 97.51 & 95.56   & \textbf{98.54}\\ \cline{2-6} 
                  & AA & 93.92   & 95.26 &  90.97  & \textbf{95.44} \\ \cline{2-6} 
                  & Kappa    & 96.96 & 96.67 & 94.05   & \textbf{98.05} \\ \hline
\multirow{3}{*}{MUUFL} & OA   & 96.47 & 96.00  & 95.10   & \textbf{97.96} \\ \cline{2-6} 
                  & AA   & 96.89  & 95.82 &  92.41   & \textbf{97.45} \\ \cline{2-6} 
                  & Kappa   & 96.25  & 95.31 & 93.44   & \textbf{97.27} \\ \hline
\bottomrule
\end{tabular}
\end{table*}

To further study each proposed module's influence, we conduct ablation studies on the three widely used datasets. To explore the effectiveness of using multi-modality, we implement a model merely based on HSI. With the sufficient information provided by other modality, we can observe a significant improvement of 7.19\%, 3.86\%, and 3.01\% in OA on Houston, Trento, and MUUFL, respectively. It demonstrates that the abundant information in different modalities is crucial to HSI classification. As shown in Fig.5, the element in the top-left corner of the fusion matrix refers to the HSI feature. We observe that only utilizing the information of one modality faces challenges in focusing on the correct part of inputs.

Nevertheless, directly fusing the information of different modalities cannot promise a significant improvement in HSI multi-modal classification. To study the influence of the proposed fusion matrix, we combine the nine elements in the fusion matrix through concatenation, replacing the fusion matrix and the feature fusion layer. Compared to the single modality, the multi-modality model using the concatenation layer only improved the metric by 1.04\%, 2.05\%, and 1.10\%, respectively, in OA on Houston, Trento, and MUUFL, which is much less significant than the improvement brought the proposed fusion matrix. The main reason is that multi-modality also introduces more noise into the model, although it contains more information. However, the proposed fusion matrix can eliminate the negative influence by adaptively adjusting the weights of different modalities. As shown in Fig. 5, the attention of each element in the fusion matrix mainly focuses on the wrong spatial position. However, after the Interconnected fusion stage, the overall attention focuses on proper spatial position, consistent with human perception.

Finally, we remove the center patch intra-modality feature encoding module to implement a model without considering neighborhood relations. We can observe a performance degradation on the model without neighborhood relations by 1.10\%, 0.79\%, and 1.49\%, respectively, in OA on Houston, Trento, and MUUFL. The fact illustrates that the contextual information of the neighborhood patches should also be included in HSI multi-modal classification. As shown in Fig.5, the four vertices of the fusion matrix refer to the fusion matrix without contextual information. The features of these four vertices have the probability of focusing on the wrong part. 

\section{Conclusions}

We proposed an Interconnected Fusion framework that considers the wealth correlations from different modalities. The contextual information between the center patch and nearby patches can be extracted to enrich the information from intra- and inter-modality. across HSI and LiDAR signals contain intra- and inter-modalities information. In addition, the correlation of intra- and inter-modalities are generated through nine different perspectives in the fusion matrix. These nine interrelated elements in the fusion matrix can complement each other and work together to eliminate bias from each view. Thus, our method can gain an in-depth understanding of the feature fusion for HSI-LiDAR signals.


\section*{Acknowledgment}
The authors would like to thank the University of Houston and the GRSS-DFC for providing the University of Houston data sets.


%





\ifCLASSOPTIONcaptionsoff
  \newpage
\fi





\bibliographystyle{IEEEtran}
\bibliography{IEEEabrv,name}

\vfill

\vfill

\end{document}